\documentclass[a4paper,conference]{IEEEtran}
\usepackage{cite}

\ifCLASSINFOpdf
   \usepackage[pdftex]{graphicx}
   \graphicspath{{images}}
\else
\fi
\begin{document}
%
\title{NL-FCOS: Improving FCOS through Non-Local Modules for Object Detection}

\author{
\IEEEauthorblockN{Lukas Pavez}
\IEEEauthorblockA{Department of Computer Science\\ University of Chile, Chile\\
email: lukas.pavez@ug.uchile.cl}
\and
 \IEEEauthorblockN{José M. Saavedra}
\IEEEauthorblockA{Universidad de los Andes, Chile \\
email:jmsaavedrar@miuandes.cl}
}

\maketitle

\begin{abstract}
During the last years, we have seen significant advances in the object detection task, mainly due to the outperforming results of convolutional neural networks. In this vein, anchor-based models have achieved the best results. However, these models require prior information about the aspect and scales of target objects, needing more hyperparameters to fit. In addition, using anchors to fit bounding boxes seems far from how our visual system does the same visual task. Instead, our visual system uses the interactions of different scene parts to semantically identify objects, called perceptual grouping. An object detection methodology closer to the natural model is anchor-free detection, where models like FCOS or Centernet have shown competitive results, but these have not yet exploited the concept of perceptual grouping. Therefore, to increase the effectiveness of anchor-free models keeping the inference time low, we propose to add non-local attention (NL modules) modules to boost the feature map of the underlying backbone. NL modules implement the perceptual grouping mechanism, allowing receptive fields to cooperate in visual representation learning. We show that non-local modules combined with an FCOS head (NL-FCOS) are practical and efficient. Thus, we establish state-of-the-art performance in clothing detection and handwritten amount recognition problems.
\end{abstract}
%
\IEEEpeerreviewmaketitle

\section{Introduction}
\label{sec:introduction}
After the explosion of deep learning \cite{alexnet},  we have seen significant advances in different computer vision tasks, not only in research but also in industrial applications. One of the tasks highly favored by these advances is object detection \cite{fasterrcnn, yolov1, efficientdet, swint,FCOS}. So far we know, anchor-based models have shown the highest performance in this task \cite{efficientdet}. An anchor is a prior, represented by a starting rectangle that a model tries to transform to fit an object in an image, predicting a tight bounding box. Anchor-based models require extra parameters like the aspect and scales of the anchors to be used, which could be hard to be defined in advance. Furthermore, anchor-based strategies are far from how our visual system does that same task naturally. Indeed, our visual system does not transform rectangles to fit objects in a scene. In contrast, it processes the whole scene taking into account the relationships between different regions to locate objects. This phenomenon is called  \emph{perceptual grouping} and it is well described in the Palmer's book \cite{book}. %
The anchor-free detection models are closer to our biological visual system, which intends to predict bounding boxes directly from the whole underlying image. In this domain, models like FCOS \cite{FCOS}, CornerNet \cite{cornernet}, CenterNet\cite{centernet} have shown competitive performance without requiring any prior rectangles. 

\begin{figure}
\centering
\includegraphics[scale = 0.4]{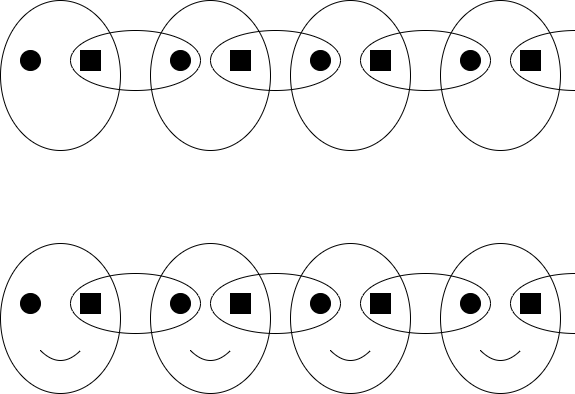}
\caption{An example of perceptual grouping \cite{book}. We can readily detect faces in the second row.}
\label{fig:grouping}
\end{figure}

Perceptual grouping studies how the various elements in a complex display are perceived as going together in one's perceptual experience \cite{book}. This phenomenon was studied by Wertheimer \cite{book} during the early past century, proposing a set of factors called \emph{the principles of grouping} that includes proximity, similarity, synchrony, among others. The interrelation of different visual components (a.k.a. visual structure) may be learned from experience, particularly during the first months of our lives. A simple example of how different parts of a scene stimulate our visual system is depicted in Figure \ref{fig:grouping}. If we look only at the first row, identifying existing groups doesn't seem very easy. However, if we examine the second row, we will rapidly note the existence of faces. It would be the consequence of our learned experience.

In computer vision, \emph{attention} is the mechanism related to perceptual grouping. It has shown to be a critical module in natural language processing \cite{attention}, and, recently, has also shown a positive impact in computer vision tasks \cite{vit, swint, polyvit}. 

Since perceptual grouping is critical in our visual system, it should also significantly impact computer vision tasks like object detection, mainly when dealing with anchor-free models. However, these models have not yet leveraged the high interrelation of different receptive fields to define the limits of the objects. Therefore, in this work, we evaluate the impact of perceptual grouping through non-local attention mechanisms \cite{nonlocal, attention} for anchor-free object detection.

The organization of this document follows the next structure. Section \ref{sec:related_work} describes the state of the art in the object detection context. Section \ref{sec:proposal} in-depth explains our proposal. Section \ref{sec:experiments} focus on the description of our experiments along with their results. Finally, Section \ref{sec:conclusions} is devoted to outline the final conclusions.

\section{Related Work}
\label{sec:related_work}
In the last years, computer vision has undergone a real boom, becoming a critical area across various industry sectors, including healthcare, agriculture, retail, manufacturing, entertainment, and others.

Object detection is one of the computer vision tasks with multiple industry applications. Its goal is to localize and classify objects in an image or video. The localization is carried out by predicting a bounding box enclosing each target object in an image. For instance, in e-commerce, searching by images is increasingly being demanded. To improve the performance in retrieval, search models decompose a catalog image and a query into semantic components using an object detection tool. 

2015 was, notably, a particular year as this marked the expansion of convolutional neural models into a vast number of computer vision tasks like object detection \cite{fasterrcnn}, segmentation \cite{fcn}, and synthesis \cite{gan}. Faster R-CNN \cite{fasterrcnn} is a milestone in the evolution of object detectors as this was the first end-to-end model using convolutional neural networks, proposing strategies that are still part of the core of current state-of-the-art methods. For instance, training bounding-box regressors guided by anchors and using a region proposal module connected to a shared backbone are some of these core components. An anchor is an a-prior knowledge a model can use to facilitate the learning process and, consequently, the convergence. Formally, an anchor is a rectangle defined by its aspect ratio and scale (area) located at each receptive field in the feature map. In this way, regressors are trained, looking for the parameters that best transform the anchors to fit the objects in the image. Although anchors seem to be a good component for training object detectors, estimating their parameters like scale and aspect could be a struggling task. Moreover, anchor-based training could reduce the generalization to unseen objects. 
In addition, as we discussed in Section \ref{sec:introduction}, the use of anchors is far from the natural behavior that, instead, is closer to what researchers call \emph{perceptual grouping}. This situation has motivated a part of the community to focus on anchor-free models like CornerNet \cite{cornernet}, or FCOS \cite{FCOS}. Thus, we will briefly describe the more famous approaches divided by their strategy to learn bounding boxes in the following lines. We have two groups, the first based on anchors and the other with anchor-free models.
\subsection{Anchor-based models}
\begin{itemize}
\item Faster-RCNN \cite{fasterrcnn}: This is a 2-stage model composed of a region proposal branch called RPN and a classification branch. Both branches share a backbone devoted to learning a visual representation map (feature map). The RPN branch tries to detect objects in a class-agnostic manner, predicting bounding boxes and an objectness score that determines the likelihood of an object. RPN introduces anchors to help it to localize objects. Candidate objects are then passed through the other branch to classify and refine the candidate bounding box with its class information.
\item YOLO (from v2): YOLO \cite{yolov1} was presented to provide a real-time model. As its name suggests (you only look once), it comprises only one stage, modeling the classification and object detection by a regression process. The original version was an anchor-free architecture, but due to the success of Faster-RCNN, rapidly, the subsequent versions incorporated an anchor-based strategy.
\item RetinaNet \cite{retinanet}: This model incorporates two critical components. The first is a feature pyramid network (FPN \cite{fpn}) to extract mutliscale representations,  and the second is the focal loss, a variant of the crossentropy overcoming the bias to the abundant easy regions appearing in an image. 
\item TridentNet \cite{tridentnet}: This is another variant of a multiscale architecture, where the multiple scales are modeled by a variety of receptive field sizes implemented through dilated convolutions (a.k.a. atrous convolutions).
\item EficientDet \cite{efficientdet}: It introduces BiFPN and a model scaling strategy. The backbone is the EfficientNet \cite{efficientnet} with multiple BiFPN modules stacked as a feature extraction network. As in FPN, each output from the final BiFPN is sent to the classification and regression heads. 
\item Panet \cite{panet}: This extends FPN with a bottom-up augmentation path to allow the low-level semantic features to be propagated to the top layers and semantic high-level to benefice bottom layers. In addition, PaNet incorporates an adaptive feature pooling module.
\end{itemize}

\subsection{Anchor-free models}
\begin{itemize}
\item CornerNet \cite{cornernet}: This anchor-free approach looks for an object bounding box by detecting a pair of keypoints, the top-left corner and the bottom-right corner of the target box. Since keypoints lack discriminative visual information, the authors propose four corner pooling modules (top-corner pooling, left-corner pooling, bottom-corner pooling, right-corner pooling) that leverage information on the borders of the target object. In addition, the network produces an embedding vector for each detected corner in such a way that corners of the same object have similar embeddings. A drawback of this model is the high prediction time due to the costly convolutional network used as a backbone (Hourglass-Net).

\item CenterNet \cite{centernet}: This is a natural extension of CornerNet, where the model additionally takes information from the center of the bounding box to validate the prediction. The authors also propose to merge each pair of corner pooling in a sequential manner, which allows the model to leverage information inside the whole bounding box instead of the borders only.

\item FCOS \cite{FCOS}: FCOS solves the object detection problem in a \textit{per-pixel} methodology, where each pixel corresponds to background or is inside an object. FCOS predicts a 4D vector with the distances from the pixel to the border of the target objects (Fig. \ref{fig:fcos_regression}.). It uses a ResNet-50 as a backbone, together with a feature pyramid network. A header is connected from each pyramid level with shared weights between the scale levels. The detection heads consist of a classification branch and a regression one. An additional computation is added for centerness at the end of one of the branches, where it shows better results at the regression branch.

\begin{figure}[!t]
\centering
\includegraphics[scale = 0.3]{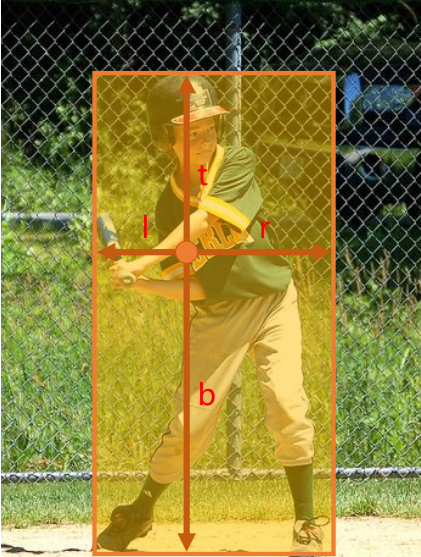}
\caption{4D regression predicted by FCOS \cite{FCOS}.}
\label{fig:fcos_regression}
\end{figure}
\end{itemize}

As we can see, the predominant backbones in the detection task are based on convolutional neural networks. Their effectiveness for extracting discriminative visual representations makes them also appropriate for other vision tasks. In contrast, attention-based models have been the favorite approaches in natural language processing \cite{attention, devlin:2019, brown:2020}, mainly due to their long-term dependency mechanism. Attention is also a critical component in our natural visual system as it allows receptive fields to interact with each other, producing the \emph{perceptual grouping} effect.

The combination of attention with convolutional models for computer vision tasks has been studied recently. Some approaches include attention as a complement of convolutional-based backbones \cite{nonlocal, hu:2020, srinivas:2021}. Others have proposed pure attention architectures for visual tasks \cite{vit, zhao:2020, swint}.

The problem with pure-attention architectures is their quadratic complexity, making them impractical to apply at high resolutions. On the other hand, convolutional models have shown to efficiently extract discriminative visual representation in multiple scales \cite{fpn, panet}. Thus, combining a convolutional backbone for feature extraction with attention modules for perceptual grouping is a natural strategy.
Although we have seen outstanding results with anchor-based approaches, the use of anchors adds an extra cost and the difficulty of defining the required parameters. Furthermore, it is far from how our natural system does the same task. Anchor-free models address this problem, but they have not gotten over the performance of anchor-based approaches yet. 

Our research indicates that the anchor-free architecture FCOS \cite{FCOS} is effective and efficient concerning others of the same type. In addition, it can be further improved by attention mechanisms. In this vein, the non-local attention mechanism  \cite{nonlocal} seems to produce the same effect of perceptual grouping as it takes into account the interrelation of different receptive fields to produce a visual representation at a specific point. In addition, the interaction between components is learned during training, simulating the learned experience discussed in Section \ref{sec:introduction}.

Therefore, this proposal presents NL-FCOS, an improved version of FCOS, that implements non-local attention mechanism after the feature map computation. Our experiments show new state-of-the-art results for clothing detection as well as for handwritten digit separation (a key component in amount recognition systems \cite{hochulli}).

\section{Non-Local FCOS}
\label{sec:proposal}
\begin{figure}[!t]
\centering
\includegraphics[scale = 0.15]{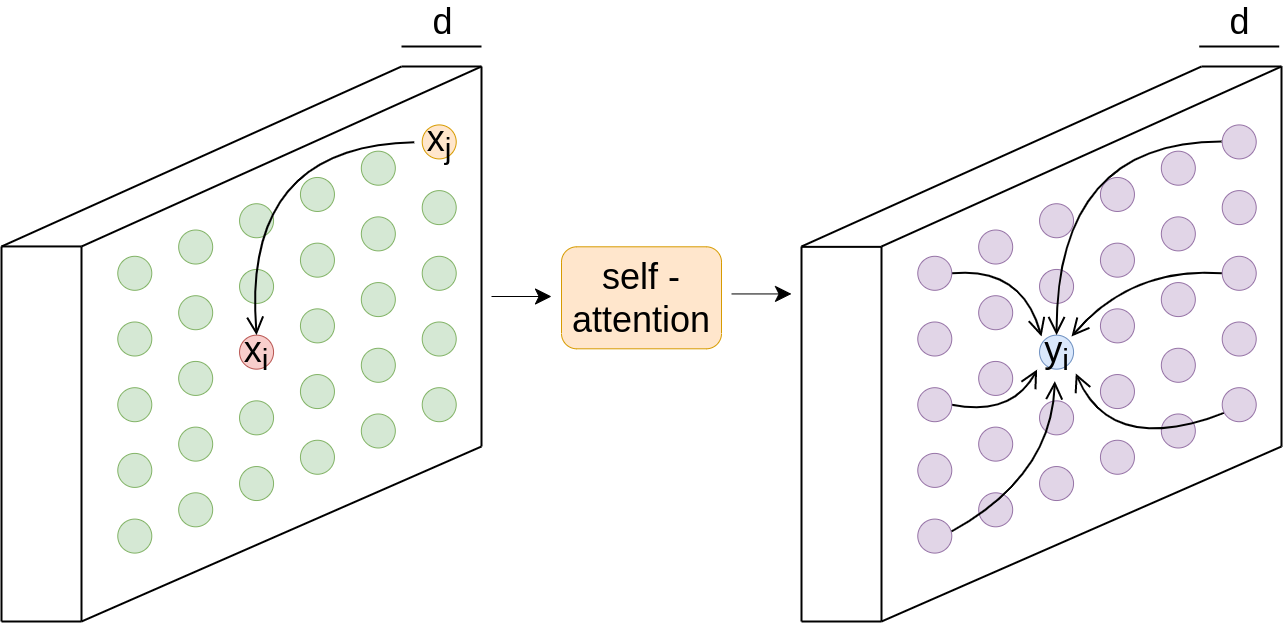}
\caption{Scheme of the attention mechanism in a non-local block in NL-FCOS.}
\label{fig:non-local}
\end{figure}

We propose a boosted FCOS, called NL-FCOS. This is an FCOS architecture connected to detection heads, which incorporate non-local attention mechanism. Our proposed attention module is based on the Non-Local Neural Networks \cite{nonlocal}, where it computes a visual representation at a position $i$ as a weighted sum of the features at all positions. This mechanism is also the core of the popular Transformers \cite{attention}. Figure \ref{fig:non-local} depicts a scheme of the non-local attention. This kind of attention mechanism implements the \emph{perceptual grouping} occurring in our visual system, where different parts of an image stimulate our system as they were one single stimulus. Since a non-local module applies a function relating each position with the others, it could also learn the identity relationship, which will produce the same result as the original model. Therefore,  the proposed NL-FCOS should perform equally or better than the original FCOS.

In Equation \ref{eq:non-local}, we formally define the non-local attention used in this proposal. The output $y_i$ is the resulting visual representation at position $i$, incorporating information from the other positions. The relationship between the representations at $i$ and $j$ is learned during training. This behavior is also connected to our visual system, where the stimuli produced by different components of a scene are learned from experience (See Fig. \ref{fig:grouping}). In Equation \ref{eq:non-local}, $\theta$, $\phi$ and $g$ are linear transformations that are directly implemented by 3x3 convolutions. Finally, the non-local transformation is then easily computed by a matrix multiplication \cite{nonlocal}

\begin{equation}
\label{eq:non-local}
    y_i = \sum_j f(\theta(x_i), \phi(x_j))g(x_j)
\end{equation}

\subsection{NL-FCOS}
We choose FCOS as the free-anchor detection architecture because it provides an excellent trade-off between efficiency and effectiveness. As we will see later, FCOS allows us to get competitive precision with a prediction time three times faster than the other anchor-free approaches. 

\begin{figure}[!t]
\centering
\includegraphics[scale = 0.3]{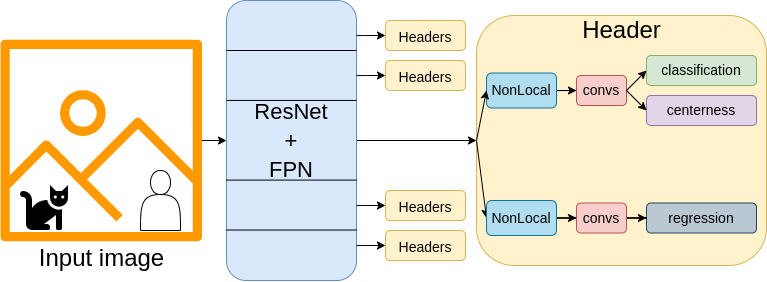}
\caption{NL-FCOS architecture.}
\label{fig:nlfcos}
\end{figure}

We modify FCOS by adding a non-local module at the beginning of each detection head after the feature extraction stage. This is because the network computes a feature map with the necessary features for the detection. The non-local block will relate each feature with the rest, adding more information to the underlying representation. Figure \ref{fig:nlfcos} displays  our proposed NL-FCOS architecture. We have also run experiments with multi-head attention, similar to what Transformer does \cite{attention}.

\subsection{Shared Non-Local Modules}
Similar to FCOS, our proposal NL-FCOS also implements a multi-scale feature extraction process using FPN \cite{fpn} and PaNet \cite{panet}. In a multiscale framework, objects closer to the observer (those appearing bigger in a scene)  require bigger receptive fields modeled by deeper layers. While objects far from the observer (those smaller) require smaller receptive fields modeled by the first layers. Through FPN or PaNet, we can model these scale variations accessing the layers that better represent the objects in a scene. In this way, the features produced by the different scales are comparable in visual representation. That is, the multiscale mechanism acts as a normalizer of objects from different scales. Thus, the classifiers and regressors in the different levels of the pyramidal representation should share their learnable parameters. 

For the non-local modules, the model needs to fit $\theta$, $\phi$, and $g$. As the visual representations extracted are comparable between the pyramid levels, it is natural that the involved parameters be shared. Therefore, in our proposal, the three transformations $\theta$, $\phi$ and $g$, required by the non-local block, are the same in all the levels. 


\subsection{NL-CornerNet}
\begin{figure}[!t]
\centering
\includegraphics[scale = 0.2]{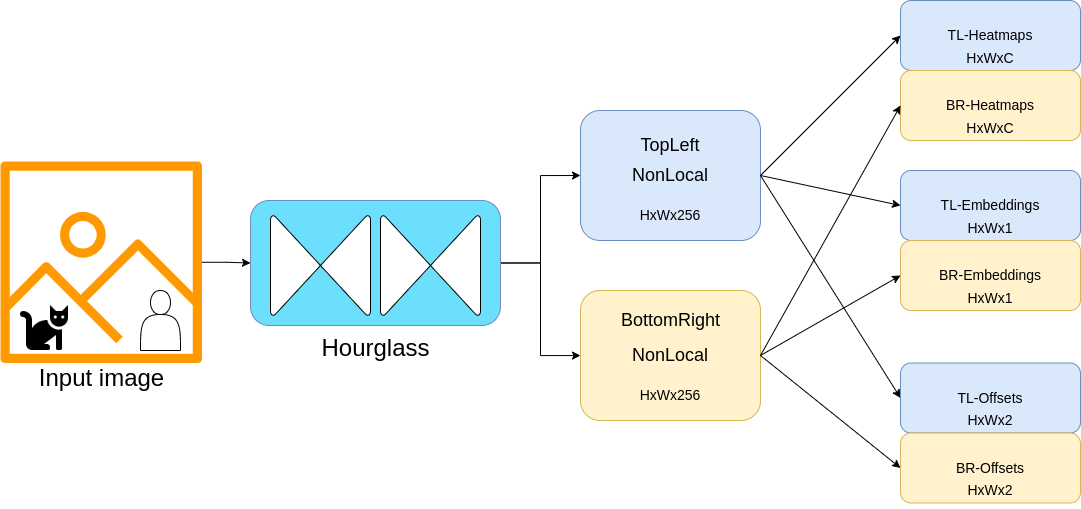}
\caption{NL-CornerNet architecture.}
\label{fig:nlcornernet}
\end{figure}

In the same manner as NL-FCOS, we also extend CornerNet \cite{cornernet} with non-local attention. To this end,  we add non-local modules after the feature extraction backbone that, in this case,  corresponds to a stacked Hourglass network. Thus, a non-local module is placed at the beginning of the prediction module, replacing the proposed corner-pooling operation. Through the non-local modules, we expect to generalize this pooling operation by adding more information from the full feature map instead of only the sides of each position as CornerNet's pooling does. Figure \ref{fig:nlcornernet} illustrates our proposed NL-CornerNet architecture.

\subsection{NL-Hybrid}
\begin{figure}[!t]
\centering
\includegraphics[scale = 0.15]{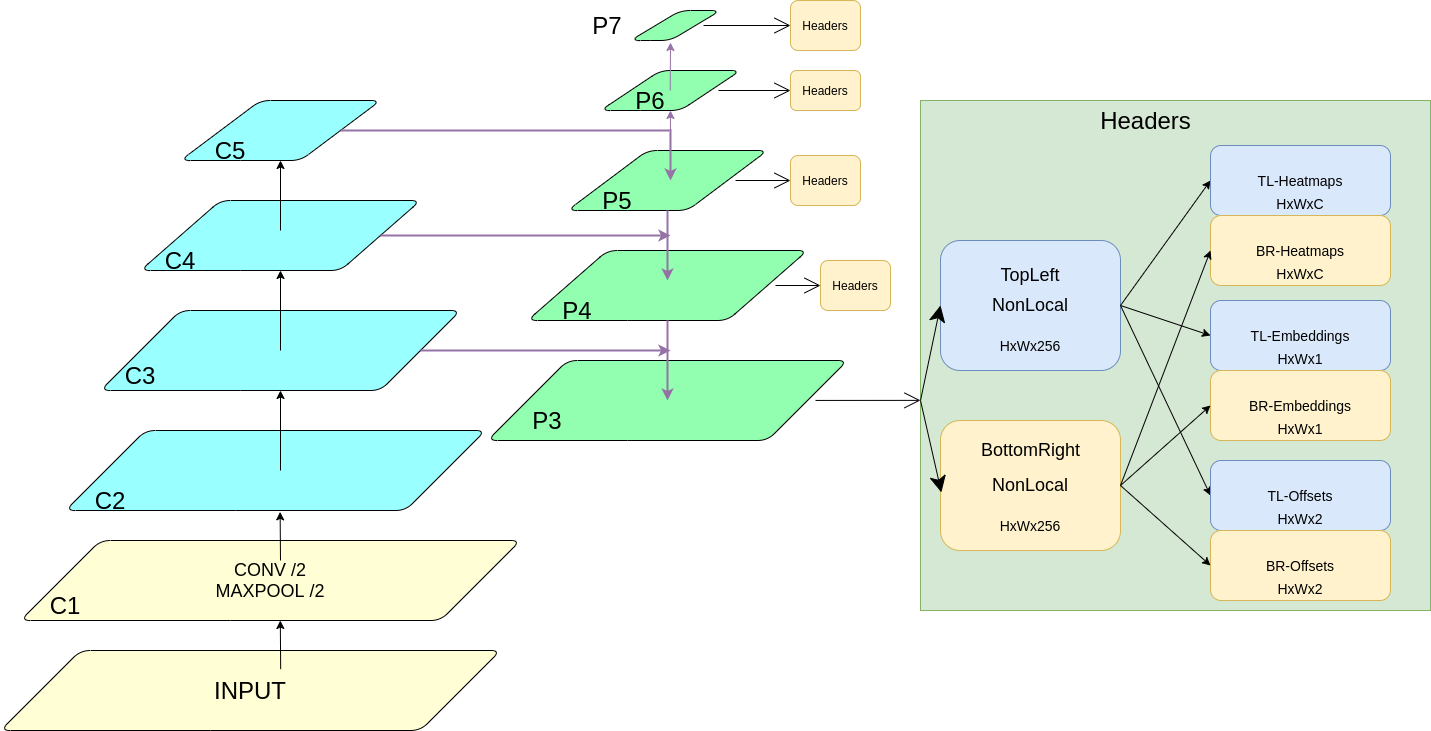}
\caption{NL-Hybrid architecture.}
\label{fig:cornerfcos}
\end{figure}

We also propose a hybrid network architecture that replaces the NL-FCOS head with the CornerNet's head. Figure \ref{fig:cornerfcos} depicts this hybrid architecture.


\section{Experimental Evaluation}
\label{sec:experiments}
\subsection{Training details}
The experiments where run with a 2-GPU server; each GPU corresponds to an Nvidia Titan RTX with 12GB. The hyper-parameter configuration follows the same used in the original proposals. We will further give more details on it.

\subsection{Datasets}
\begin{figure}[!t]
\centering
\includegraphics[scale = 0.30]{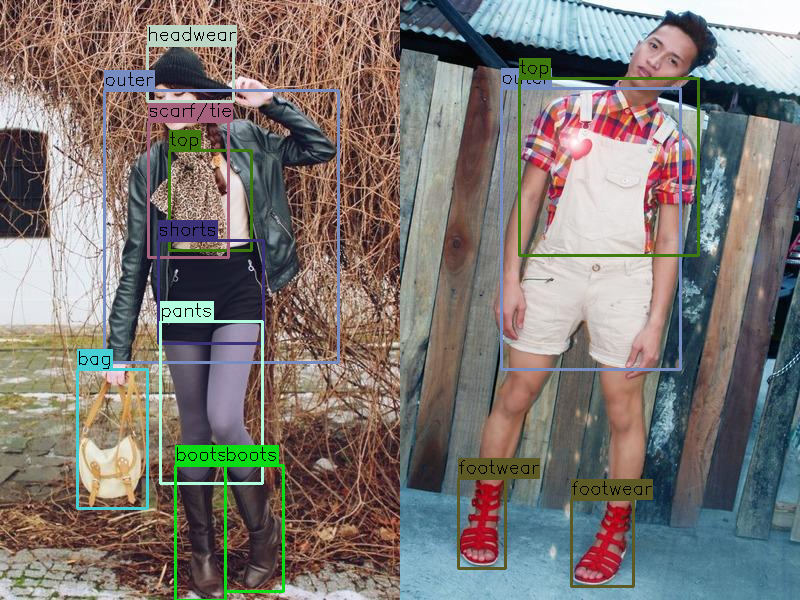}
\caption{Examples of annotated bounding boxes in the Modanet dataset.}
\label{fig:modanet_example}
\end{figure}

We run experiments with two datasets, both related to industrial applications. The first one is Modanet \cite{modanet},  a clothing dataset with around 52000 images containing boxes enclosing objects from 13 different apparel classes. We divide this dataset into training and testing sets. The first one with 42000 images and 214792 boxes. For testing, we have 10000 images and 50907 boxes. An example of an image from the Modanet dataset is presented in Figure \ref{fig:modanet_example}.

The second dataset is related to the handwritten amount recognition task. Recently, Hochulli et al. \cite{hochulli} have shown outstanding effectiveness in amount recognition using a detection-based approach to separate the handwriting digits. Here, we use the two check amount datasets proposed by Diem et al. \cite{hdsrc}, which were then bounding-box annotated in the work of Hochulli et al. \cite{hochulli}. These are CAR-A and CAR-B, which are described below:
\begin{itemize}
    \item CAR-A: This consists of 1500 images with 6744 boxes for training; and 509 images with 2328 boxes for testing.
    \item CAR-B: This consists of 1000 images with 5621 boxes for training; and 500 images with 2812 boxes for testing. 
\end{itemize}

\begin{figure}[!t]
\centering
\includegraphics[scale = 0.70]{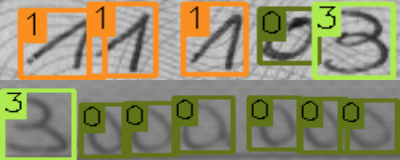}
\caption{An example of annotated bounding boxes in the CAR-A (top) and CAR-B (bottom) datasets.}
\label{fig:car_example}
\end{figure}

The main difference between these two datasets is the resolution used for capturing the check images. CAR-A images were captured with a higher resolution, producing images with better quality. An example image of each dataset is illustrated in Figure \ref{fig:car_example},  along with their annotated bounding boxes.


\subsection{NL-FCOS}
The network is trained with the SGD optimizer for about 90000 iterations having a batch size of 8 images. The learning rate is initially set to 0.01 and is then reduced by a factor of 10 by the iteration 60K. The backbone is previously pretrained on the ImageNet dataset \cite{imagenet}. 

As shown in Table \ref{table:fcos_ap}, we run experiments aiming to decide the branch in the prediction block where the non-local module should be located. We evaluated three branches: classification (NL-FCOS-cls), regression (NL-FCOS-reg) and both (NL-FCOS). Our results show that the non-local block increases the performance regardless the branch where it is located. We see that putting it in one branch or both produces similar results, but they are superior to the original model without attention. Our results also present a new state-of-the-art in the Modanet dataset, where the best reported anchor-based model is TridentNet, achieving an AP of 54.0 \cite{simon}. We also run experiments with a multihead attention (FCOS-transformer), running two non-local attention blocks. Here, the results are lower than using only one attention module by almost $1\%$. It is possible that the model requires more iterations to adjust a greater number of weights generated by the second non-local module that runs parallel to the other. 

In regard to inference time, as shown in Table \ref{table:fcos_time}, the difference caused by adding the non-local module is marginal (this adds 2ms). 

\begin{table}[!t]
\centering
\caption{Comparing FCOS and NL-FCOS in the Modanet dataset. The first row shows the AP for the best anchor-based model in this dataset \cite{simon}.}
\label{table:fcos_ap}
\begin{tabular}{|c|c|c|c|}
\hline
\textbf{Network} & \textbf{$AP$}   & \textbf{$AP^{50}$} & \textbf{$AP^{75}$}  \\ \hline
TridentNet \cite{simon} & 54.0 & 77.5 & 62.2 \\ \hline
FCOS & 59.6 & 79.3 & 66.4 \\ \hline
NL-FCOS & 62.0 & 82.3 & 69.2 \\ \hline
NL-FCOS-cls  & 62.1 & 82.2 & 68.9 \\ \hline
NL-FCOS-reg & 62.1 & 82.0 & 68.4 \\ \hline
FCOS-transformer & 60.3 & 81.3 & 67.0 \\ \hline
\end{tabular}
\end{table}

\begin{table}[!t]
\centering
\caption{Average inference time for FCOS and NL-FCOS.}
\label{table:fcos_time}
\begin{tabular}{|c|c|}
\hline
\textbf{Network} & \textbf{Time [ms]}   \\ \hline
FCOS & 33.0 \\ \hline
NL-FCOS & 35.0 \\ \hline
NL-FCOS-cls & 35.0 \\ \hline
NL-FCOS-reg & 35.0 \\ \hline
\end{tabular}
\end{table}

We also run experiments with CAR-A and CAR-B datasets. First, we train our model independently in each dataset. We observe that in both datasets, NL-FCOS produces a slight improvement. We also trained our model combining both sets, obtaining higher performance (see Table \ref{table:fcos_montos_ab}). After applying the non-local module, we observe that the mean precision increases to $99.0$, which is $0.2\%$ better than the simple FCOS. We also obtain results comparable to the state-of-the-art results obtained by YOLO \cite{david}, where we improve these results in the dataset CAR-B.

\begin{table}[!t]
\centering
\caption{AP values obtained in CAR-A y CAR-B datasets.}
\label{table:fcos_montos_ab}
\begin{tabular}{|c|c|c|c|}
\hline
\textbf{Network} & \textbf{CAR-A} & \textbf{CAR-B} & \textbf{CAR-A + CAR-B} \\ \hline
YOLO \cite{david}     & 96.8           & 96.5  & -         \\ \hline
FCOS         & 96.3           & 98.4           & 98.8  \\ \hline
NL-FCOS      & 96.5           & 98.5           & 99.0 \\ \hline
\end{tabular}
\end{table}

\begin{figure}[!t]
\centering
\includegraphics[scale = 0.25]{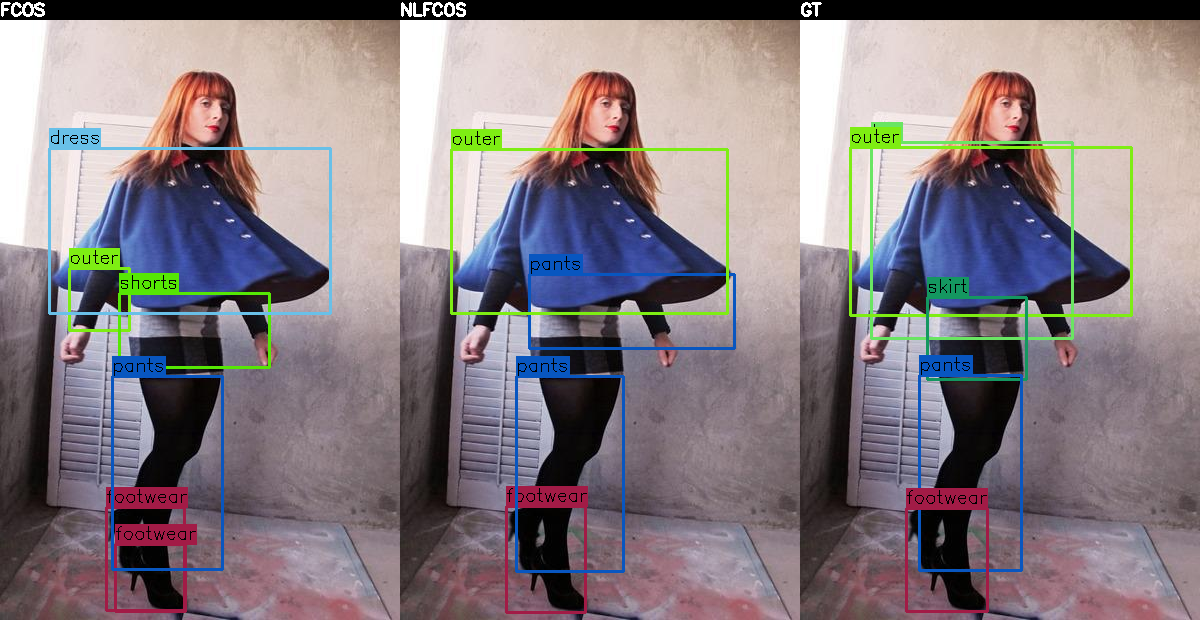}
\caption{Detection comparison between FCOS and NL-FCOS for the Modanet dataset.}
\label{fig:modanet_detection}
\end{figure}

\begin{figure}[!t]
\centering
\includegraphics[scale = 0.6]{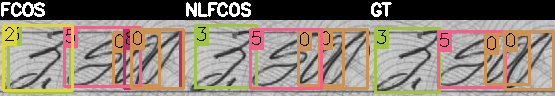}
\caption{Detection comparison between FCOS and NL-FCOS for the check ammount dataset.}
\label{fig:montos_detection}
\end{figure}

The handwriting digit detection is a critical component in the amount recognition task \cite{hochulli}. To measure how well our proposal behaves in this context,  we add two metrics: accuracy and similarity. The accuracy increments when the predicted amount is the same as expected. The similarity indicates how close was the predicted amount to the desired amount (for example, the similarity between $2871$ and $28071$ is $0.89$). This helps to identify in which case we are detecting more or fewer digits of the number. The results are shown in Table \ref{table:fcos_recognition}, where FCOS achieves a low accuracy but a high similarity. Here in all cases, using the non-local modules increases the results in both datasets. We can see some results in Figures   \ref{fig:modanet_detection} and \ref{fig:montos_detection}.

\begin{table}[!t]
\centering
\caption{Accuracy (acc) and similarity (sim) in the amount recognition task for CAR-A and CAR-B.}
\label{table:fcos_recognition}
\begin{tabular}{|c|cc|cc|cc|}
\hline
                 & \multicolumn{2}{c|}{\textbf{CAR-A}} & \multicolumn{2}{c|}{\textbf{CAR-B}} & \multicolumn{2}{c|}{\textbf{CAR-A + CAR-B}} \\ \hline
\textbf{Network} & \multicolumn{1}{c|}{acc}    & sim   & \multicolumn{1}{c|}{acc}    & sim   & \multicolumn{1}{c|}{acc}        & sim       \\ \hline
FCOS             & \multicolumn{1}{c|}{48.7}   & 89.5  & \multicolumn{1}{c|}{72.4}   & 95.3  & \multicolumn{1}{c|}{82.6}       & 96.8      \\ \hline
NL-FCOS          & \multicolumn{1}{c|}{74.3}   & 95.1  & \multicolumn{1}{c|}{75.4}   & 96.3  & \multicolumn{1}{c|}{83.2}       & 96.8      \\ \hline
\end{tabular}
\end{table}


\subsection{NL-CornerNet}
The network is trained with Adam optimizer for 250.000 iterations with a batch of 4 images. The initial learning rate is $0.25 x 10^{-4}$. Due to memory limitations, we reduce the size of the input of the prediction module from $128\times128$ to $64\times64$ in all our experiments. 

Analyzing the results shown in Table \ref{table:cornernet_ap}, we note that the non-Local module reduces the AP value. This goes contrary to expected and can have different causes. A cause of this behavior could be the Hourglass network that already combines all the features, and doing it again in the prediction module may be damaging to the general result. Another hypothesis is that the network needs more iterations to adapt the weights of non-local modules. We are inclined to the second reason since a non-local module could learn identity attention, keeping the original performance.

A strong limitation of CornerNet is about its inference time (Table \ref{table:cornernet_time}). This network is much slower than FCOS (almost three times slower), and applying the non-local module adds about 17\% of the original network time, which is not that much considering that the network is already slow. Therefore, our proposal NL-FCOS is still a good choice in real-time scenarios. 

\begin{table}[!t]
\centering
\caption{Experiments with CornerNet and Non-Local module, dataset ModaNet.}
\label{table:cornernet_ap}
\begin{tabular}{|c|c|c|c|}
\hline
\textbf{Network} & $AP$   & $AP^{50}$ & $AP^{75}$  \\ \hline
CornerNet & 67.2 & 82.9 & 73.5 \\ \hline
NL-CornerNet & 61.8 & 79.4 & 68.0 \\ \hline
\end{tabular}
\end{table}

\begin{table}[!t]
\centering
\caption{Average inference time for CornerNet experiments.}
\label{table:cornernet_time}
\begin{tabular}{|c|c|}
\hline
\textbf{Network} & \textbf{Time [ms]}   \\ \hline
CornerNet & 100.0 \\ \hline 
NL-CornerNet & 117.5 \\ \hline
\end{tabular}
\end{table}

\subsection{NL-Hybrid}
The network is trained with SGD for about 250.000 iterations with a batch size of 8 images. The learning rate is initially set to 0.001, and it is then reduced by a factor of 10 on the iteration 200K. The backbone is pretrained on ImageNet \cite{imagenet}. 

The experiments with this architecture are divided in two parts. First, we apply the prediction module to only one level of the Feature Pyramid (level P3). The results in these experiments (Table \ref{table:cornerfcos_single}) are lower than the results obtained with FCOS and CornerNet alone, but we note again the goodness of using a non-local module as it gives better results than not using it.  In the second experiment, we apply multiscale detection using the Feature Pyramid Network. In this case, adding the non-local module does not impact the performance. 

\begin{table}[!t]
\centering
\caption{Results with NL-Hybrid, where (single) express a monoscale detection on P3, and multi is a classical multiescale detection using FPN.}
\label{table:cornerfcos_single}
\begin{tabular}{|c|c|c|c|}
\hline
\textbf{Network} & $AP$   & $AP^{50}$ & $AP^{75}$  \\ \hline
Hybrid (single) & 47.7 & 66.6 & 53.0 \\ \hline
NL-Hybrid (single)& 52.2 & 71.2 & 58.2  \\ \hline
NL-Hybrid (multi)  & 43.3 & 68.4 & 46.4  \\ \hline
Hybrid (multi) &  43.3 & 68.2 & 46.5  \\ \hline
\end{tabular}
\end{table}

\section{Conclusions}
\label{sec:conclusions}
We present NL-FCOS, an improved version of FCOS network, augmented with a non-local attention mechanism in the detection head. As shown in our experimental results, the application of attention favorably affects FCOS, increasing the detection precision by 3\%, and keeping inference time low. This is not only in the clothing dataset but also in the handwritten amounts dataset. We also explore multihead attention as Transformers, but our results do not show any improvement. 

Our experimental datasets are close to industrial applications, Modanet is used in fashion e-commerce, while CAR-A and CAR-B are used in check amount recognition. Our proposal shows better results than the corresponding best reported anchor-based method in both cases.







%



\bibliographystyle{ieeetr} 
\bibliography{references}

\end{document}